\documentclass[runningheads]{llncs}
\usepackage[T1]{fontenc}
\usepackage{graphicx}
\usepackage{enumitem}
\usepackage{amsmath,amssymb,bm}
\usepackage{booktabs}
\usepackage{graphicx}
\usepackage{hyperref}
\usepackage{listings}
\usepackage{subcaption}
\usepackage{xcolor}
\usepackage{float}
\usepackage{enumitem}
\usepackage{pdflscape} 
\usepackage{pdfpages}
\usepackage{amsmath,amssymb,booktabs,enumitem}

\hypersetup{colorlinks=true,linkcolor=blue,citecolor=blue,urlcolor=blue}

\begin{document}
\title{MultiSense-Pneumo: A Multimodal Learning Framework for Pneumonia Screening in Resource-Constrained Settings}
\titlerunning{MultiSense-Pneumo for Pneumonia Screening}
%
%\titlerunning{Abbreviated paper title}
% If the paper title is too long for the running head, you can set
% an abbreviated paper title here
%
\author{Dineth Jayakody\inst{1}\orcidID{0009-0009-8926-7476} \and
Pasindu Thenahandi\inst{1}\orcidID{0009-0009-7009-3736} \and
Chameli Dommanige\inst{1}\orcidID{0009-0005-6179-5403}}
\authorrunning{Jayakody et al.}
% First names are abbreviated in the running head.
% If there are more than two authors, 'et al.' is used.
%
\institute{Department of Computer Science, Old Dominion University, VA, USA}
\maketitle              % typeset the header of the contribution
\begin{abstract}
Pneumonia remains a leading global cause of morbidity and mortality, particularly in low-resource settings where access to imaging, laboratory testing, and specialist care is limited. Clinical assessment relies on heterogeneous evidence, including symptoms, respiratory patterns, spoken descriptions, and chest imaging, making frontline screening inherently multimodal. However, many existing computational approaches remain unimodal and focus primarily on radiographs. In this work, we present \textbf{MultiSense-Pneumo}, a multimodal research prototype for pneumonia-oriented screening and triage support that integrates structured symptom descriptors, cough audio, spoken language, and chest radiographs. The system combines deterministic symptom triage, LightGBM-based acoustic classification, domain-adversarial radiograph analysis using ResNet-18, transformer-based speech recognition, and an interpretable late-fusion operator. Each modality is transformed into a normalized concern signal and aggregated into a unified screening estimate. The fusion weights are hand-specified and are treated as heuristic, interpretable parameters rather than learned or clinically optimized values. MultiSense-Pneumo is implemented with offline execution in mind on standard laptop-class hardware, but it is not presented as a deployment-validated or clinically validated diagnostic system. Experimental results demonstrate strong component-level performance of the radiograph pathway under synthetic domain shifts, while also highlighting important limitations, especially reduced abnormal-class recall for cough acoustics and the absence of paired end-to-end multimodal patient evaluation. MultiSense-Pneumo is therefore intended as a framework and component-level prototype for screening and triage research.

\end{abstract}
\section{Introduction}
\label{sec:intro}

Pneumonia remains one of the most important infectious causes of illness and death worldwide and continues to impose a disproportionate burden on children under five years of age, older adults, and medically vulnerable populations. Recent global burden estimates indicate that pneumonia caused approximately 2.5 million deaths worldwide in 2023, representing a marked increase relative to 2021. Children younger than five years accounted for roughly 600{,}000 of these deaths, while adults aged 70 years and older accounted for approximately 1.2 million; together, these two age groups represented nearly 70\% of global pneumonia mortality \cite{gbd2023_pneumonia,springer_pneumonia_burden_post}. The burden is also highly uneven geographically: Sub-Saharan Africa together with South and East Asia and the Pacific accounted for approximately 68\% of pneumonia deaths in 2023, underscoring persistent inequities in prevention, timely diagnosis, and access to life-saving interventions such as antibiotics and therapeutic oxygen \cite{springer_pneumonia_burden_post}. WHO further reports that pneumonia accounted for 14\% of all deaths among children under five in 2019, causing 740{,}180 child deaths, while UNICEF continues to describe pneumonia as the leading infectious cause of death in young children globally \cite{who_pneumonia_children,unicef_pneumonia_stats}. More broadly, pneumonia remains a clinically heterogeneous syndrome with substantial variation in etiology, severity, and presentation across age groups and care settings \cite{lim2022pneumonia}.

From a clinical standpoint, pneumonia assessment is intrinsically multimodal. Physicians rarely rely on a single source of evidence; instead, they integrate symptom history, respiratory complaints, physical signs, verbal observations from patients or caregivers, and, when available, radiographic evidence of pulmonary involvement \cite{lim2022pneumonia,who_pneumonia_topic}. This layered reasoning process is especially relevant in community and primary-care contexts, where patients are often first encountered by non-specialist providers or community health workers rather than by radiologists or pulmonologists. WHO’s Integrated Management of Childhood Illness (IMCI) framework was developed precisely to strengthen frontline assessment and management of common childhood conditions, including pneumonia, in primary-care and community-facing settings \cite{who_imci}. At the same time, evidence suggests that frontline workers in low- and middle-income settings may struggle with difficult pneumonia-related signs such as chest indrawing, and respiratory-rate-based screening can be operationally challenging even for trained staff \cite{jogh_chest_indrawing_review,baker2018rr}. These constraints make structured computational support particularly relevant for early-stage screening and referral.

The diagnostic bottlenecks are not limited to human expertise. Access to imaging and other diagnostic tools remains uneven, especially at the primary-care level. Recent analyses have emphasized that diagnostic imaging availability is highly constrained in many low-resource health systems, while pneumonia care pathways often depend on incomplete or delayed access to confirmatory testing \cite{nafade2024diagnostic_imaging,who_pneumonia_children}. In addition, delayed care-seeking remains a major barrier: the Springer Nature synthesis reports that 44\% of children worldwide with pneumonia symptoms are not taken to a qualified healthcare provider, and that in Sub-Saharan Africa only about 45\% of children with acute respiratory infection symptoms receive care, with even lower rates in rural areas \cite{springer_pneumonia_burden_post}. Likewise, the availability of pulse oximetry and oxygen remains a central determinant of survival in severe pneumonia, particularly where hypoxemia detection is otherwise delayed or absent \cite{cilloniz2021pulseox}. Consequently, many real-world screening decisions must be made under partial information: symptoms may be available before imaging, verbal descriptions may substitute for structured histories, and resource availability may determine whether radiography, oxygen support, or escalation can be pursued at all.

These realities motivate a computational view of pneumonia screening as an inherently \emph{multimodal inference problem}. The input channels involved are statistically heterogeneous: symptom descriptions are sparse and linguistic, cough recordings are short-duration acoustic waveforms, spoken observations are temporally structured language signals, and chest radiographs are high-dimensional medical images subject to acquisition variability and domain shift. Yet many computational systems proposed for pneumonia analysis remain strongly unimodal, typically emphasizing radiographs alone \cite{ginsburg2023ai_pneumonia}. While radiograph-based models are important, a purely imaging-centered view does not capture the broader structure of frontline clinical reasoning, especially in settings where imaging may be absent, delayed, or difficult to interpret.

MultiSense-Pneumo is designed in response to this gap. The framework integrates four complementary evidence channels, namely structured symptom descriptors, monaural cough audio, spoken natural language, and chest radiographs, within a unified screening architecture. Its aim is not to replace clinical diagnosis, but to explore how computational tools could support \textbf{screening} and \textbf{triage}: identifying individuals who may warrant closer evaluation, follow-up imaging, confirmatory testing, or urgent referral. This framing is important for community-facing and humanitarian settings, where partially available evidence is common and offline execution may be necessary. In the present work, however, these are design goals rather than deployment-validated claims; clinical usefulness would require prospective evaluation with paired multimodal data and human users.

From a machine learning perspective, pneumonia-oriented multimodal screening introduces several nontrivial challenges. Different modalities exhibit different failure modes: text may be incomplete or ambiguous, cough acoustics may be noisy and weakly discriminative, speech transcription may introduce recognition errors, and imaging models may be sensitive to distribution shifts arising from blur, contrast changes, scanner differences, and acquisition artifacts. In addition, the system must remain interpretable. In constrained deployment environments, users must be able to inspect how each modality contributes to the final estimate rather than relying on an opaque end-to-end score.

In this work, we present MultiSense-Pneumo as a modular and interpretable multimodal framework for pneumonia-oriented screening support. Each modality is mapped to a bounded scalar signal representing pneumonia-related concern, and these signals are combined through an explicit fusion operator. This design improves traceability, permits ablation by modality, and provides a computational analogue of evidence aggregation in early clinical reasoning. The contribution of this paper is therefore not a claim of clinical validation, but a technically grounded description of a multimodal research prototype together with quantitative evidence characterizing both its strengths and its current limitations.

\section{System Overview}
\label{sec}

MultiSense-Pneumo is organized as a multimodal screening framework composed of four primary clinical-information pathways:
\begin{enumerate}[leftmargin=*]
\item a deterministic symptom-triage pathway,
\item an acoustic cough-analysis pathway,
\item a speech-transcription pathway, and
\item a chest-radiograph classification pathway.
\end{enumerate}
Each pathway transforms its raw input into a bounded scalar representation of pneumonia-related concern. These modality-specific signals are subsequently integrated by an interpretable fusion mechanism to produce a unified screening score and an ordinal risk band.

Conceptually, the end-to-end computation can be summarized as
\[
u \;\mapsto\;
\Big(
\hat{s}^{\mathrm{sym}},
\hat{s}^{\mathrm{cgh}},
\hat{s}^{\mathrm{sp}},
\hat{s}^{\mathrm{img}}
\Big)
\;\mapsto\;
S,
\]
where $u$ denotes the set of user-provided inputs, $\hat{s}^{\mathrm{sym}}$ is the structured symptom signal, $\hat{s}^{\mathrm{cgh}}$ is the cough-acoustic signal, $\hat{s}^{\mathrm{sp}}$ is the speech-derived symptom signal, $\hat{s}^{\mathrm{img}}$ is the radiographic signal, and $S \in [0,1]$ is the final multimodal screening score.

The framework is modular in the sense that each modality can be studied independently, replaced by a stronger component, or omitted when unavailable. It is interpretable in the sense that each intermediate signal is explicitly defined and can be inspected separately from the final score. These two design choices are important for research use, because they permit systematic analysis of which modalities dominate performance, how each component behaves under perturbation, and where failure modes arise.

\section{System Architecture}
\label{sec:architecture}

The architecture of MultiSense-Pneumo is designed to reflect real-world pneumonia assessment, where evidence is often collected from multiple partially independent sources and interpreted under incomplete information. Rather than relying on a single monolithic predictor, the system is formulated as a \emph{stacked multimodal screening framework} that processes heterogeneous inputs and maps them into a common risk space for aggregation. The architecture should be interpreted as a modular research prototype rather than as a clinically validated end-to-end diagnostic system.

\begin{figure}[htbp]
\centering
\includegraphics[width=\textwidth]{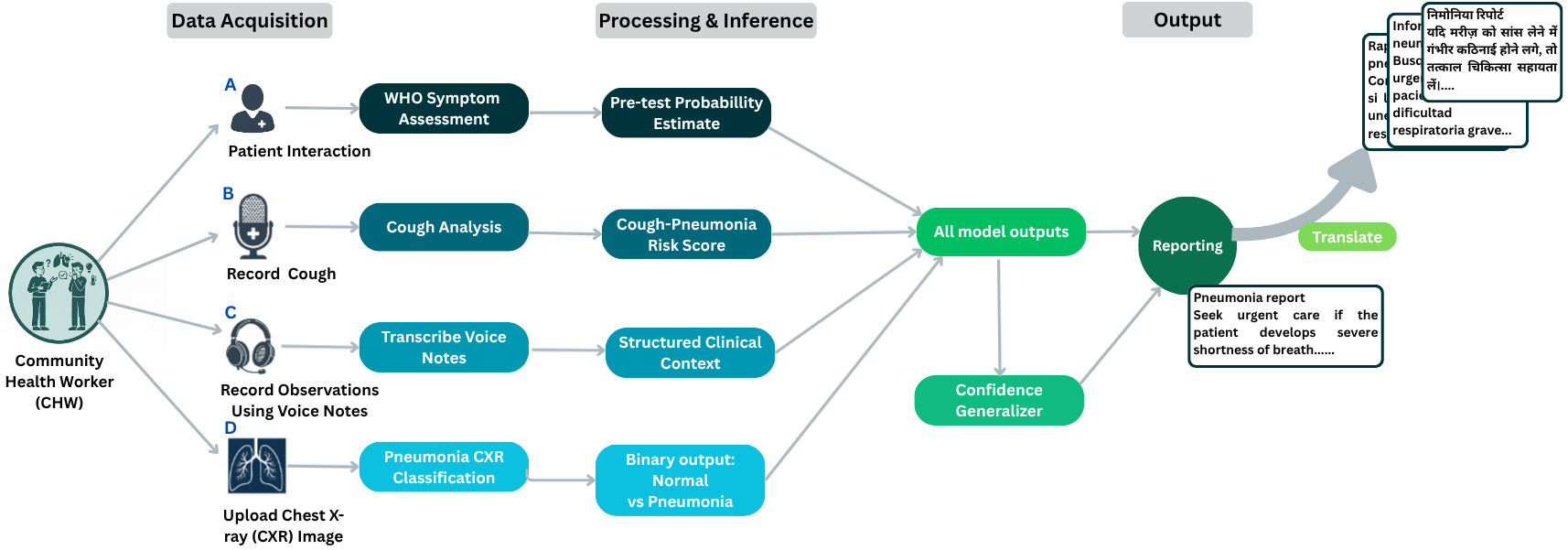}
\caption{Schematic overview of the MultiSense-Pneumo multimodal architecture. Each modality-specific branch produces a normalized risk signal, which is subsequently integrated through an interpretable fusion operator to generate a unified pneumonia-oriented screening estimate. In the current study, this fused estimate is a heuristic prototype output rather than a clinically validated diagnostic score.}
\label{fig:architecture}
\end{figure}

The framework consists of four modality-specific evidence pathways, followed by fusion and optional reporting (Fig.~\ref{fig:architecture}):

\begin{itemize}[leftmargin=*]

\item \textbf{Structured symptom analysis:} Questionnaire responses or constrained symptom descriptions are transformed into a bounded clinical concern signal using deterministic, guideline-inspired logic. This component captures primary indicators such as cough, fever, and respiratory difficulty while enforcing safety through rule-based escalation constraints.

\item \textbf{Cough acoustic classification:} Cough recordings are converted into time--frequency and low-level acoustic descriptors, from which a gradient-boosting classifier estimates an abnormal-cough probability. Because this pathway is affected by data imbalance and recording variability, its output is treated as an auxiliary signal rather than as a standalone screening decision.

\item \textbf{Speech transcription and secondary text analysis:} Spoken input is transcribed using a pretrained automatic speech recognition (ASR) model. The resulting text is analyzed for symptom-related cues, providing an auxiliary semantic signal that complements structured symptom entry.

\item \textbf{Chest radiograph classification:} Chest X-ray images are processed using a convolutional neural network (ResNet-18 backbone) augmented with domain-adversarial training (DANN). This branch provides the strongest component-level quantitative evidence in the present study and is evaluated under clean, blur, noise, and contrast-shifted conditions.

\item \textbf{Multimodal fusion:} The outputs from all available modality-specific branches are normalized and combined through an explicit weighted operator. The fusion weights are hand-specified for interpretability and should be understood as a heuristic design choice, not as learned or clinically optimized parameters.

\item \textbf{Narrative report generation:} A structured, human-readable summary is generated using a medical-domain language model (MedGemma), conditioned on the fusion score and intermediate modality-specific signals. This module is intended only to summarize upstream evidence for research use. It is not evaluated as a clinically safe report generator and must not be interpreted as producing verified diagnostic text.

\end{itemize}

Each pathway produces an intermediate scalar representation of pneumonia-related concern. These signals are aligned into a shared scale and aggregated to support transparent inspection of how each modality contributes to the final screening estimate. Because the current experiments do not use a fully paired multimodal patient dataset, the evidence in this paper supports component-level performance and framework feasibility, not clinical validation of the complete fused pipeline. Nevertheless, the proposed architecture is designed for real-world screening encounters in which symptoms, cough audio, speech descriptions, and chest radiographs are collected from the same patient. Under such paired-data conditions, the modality-specific signals and fusion operator can be applied directly to patient-aligned inputs. Thus, the current limitation is the lack of paired multimodal validation data, rather than a structural incompatibility of the framework with real-world multimodal use. Future work will evaluate the complete fused pipeline on clinically paired cohorts. 

\section{Methodology}
\label{sec:methods}

The proposed MultiSense-Pneumo framework adopts a modular, multimodal design that integrates heterogeneous clinical signals into a unified screening estimate. Each modality, including structured symptoms, cough acoustics, chest radiographs, and speech-derived information, is processed through a dedicated pathway tailored to its data characteristics. These pathways produce normalized signals that are subsequently combined through an interpretable fusion mechanism. The overall design emphasizes robustness to missing or noisy inputs, computational efficiency for deployment in resource-constrained settings, and transparency to support clinical interpretability.

\subsection{Structured symptom triage}
\label{sec:structured_symptom_triage}

\begin{figure}[htbp]
\centering
\includegraphics[width=1.0\linewidth]{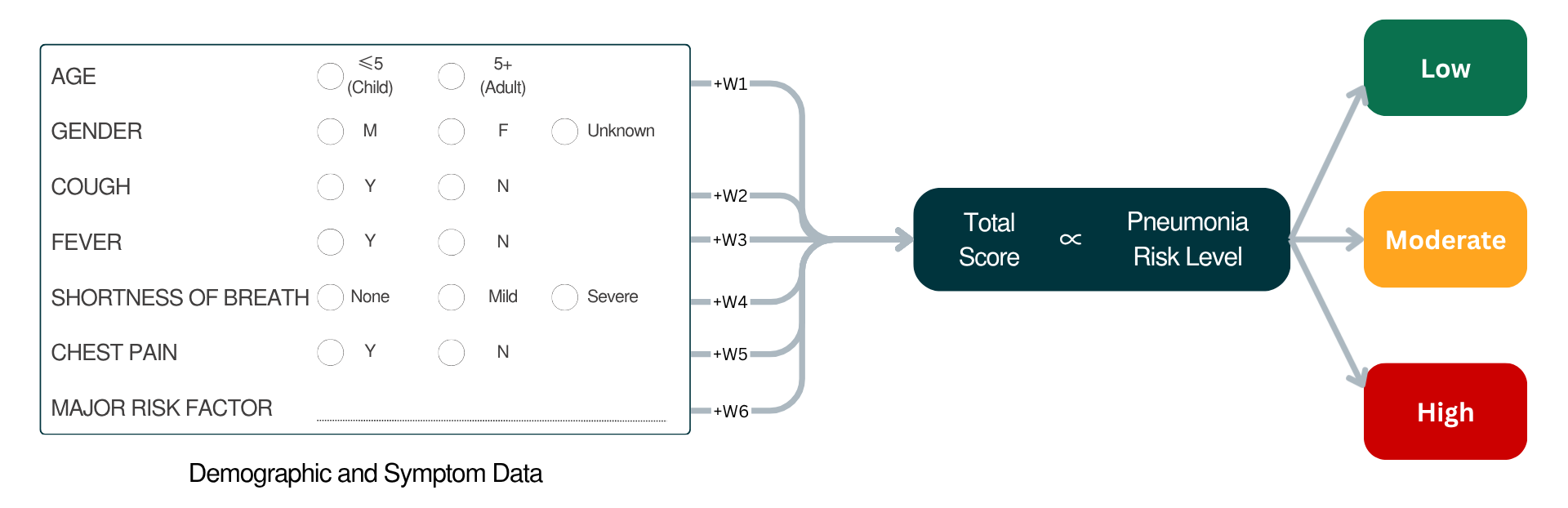}
\caption{Structured symptom triage module based on guideline-inspired assessment. User-provided inputs, including demographic attributes (age, gender), primary symptoms (cough, fever, shortness of breath, chest pain), and major risk factors, are mapped to weighted signals. These signals are aggregated into a total score, which is subsequently translated into discrete pneumonia risk levels (low, moderate, high). The design ensures interpretability and deterministic behavior, aligning with standardized clinical screening practices.}
\label{fig:symptom_triage}
\end{figure}

The symptom pathway is a deterministic triage submodule (Fig.\ref{fig:symptom_triage}) that maps structured screening answers (cough or difficult breathing, fever or chills, shortness of breath graded as none, mild, or severe, chest pain or confusion, major risk factors, and for children under five, chest indrawing or inability to drink or feed) into a discrete risk label for pneumonia-related concern. The intent is to mirror early, history-first screening when imaging is delayed or unavailable.

Certain responses bypass the additive score and force an \textsc{URGENT} band:
severe shortness of breath; chest pain or confusion; and, in the pediatric
group, chest indrawing or unable to drink/feed. These act as hard escalation
rules without invoking a probabilistic model.

Otherwise, let $\mathbb{I}_1,\ldots,\mathbb{I}_4\in\{0,1\}$ indicate,
respectively, cough/difficult breathing, fever/chills, mild (not severe)
shortness of breath, and presence of a major risk factor. With fixed weights
$w_1=2$, $w_2=1$, $w_3=2$, $w_4=1$, define the symptom score
\begin{equation}
  s_{\mathrm{sym}} \;=\; \sum_{k=1}^{4} w_k\,\mathbb{I}_k
  \;\in\; \{0,1,\ldots,6\}.
\end{equation}
Risk bands are obtained by thresholding: $s_{\mathrm{sym}}\ge 5$ yields
\textsc{HIGH}, $3\le s_{\mathrm{sym}}\le 4$ yields \textsc{MODERATE}, and
$s_{\mathrm{sym}}\le 2$ yields \textsc{LOW}. Equivalently, a normalized summary
$\hat{s}_{\mathrm{sym}}=s_{\mathrm{sym}}/6$ lies in $[0,1]$ and preserves the
same ordering. The design is transparent and auditable; its limitation is
dependence on the chosen symptom set and weights, and on structured inputs rather
than unrestricted free text in the rule layer itself.

\subsection{Cough acoustic classification}
\label{sec:cough_acoustic_classification}

\begin{figure}[htbp]
  \centering
  \includegraphics[width=1.0\linewidth]{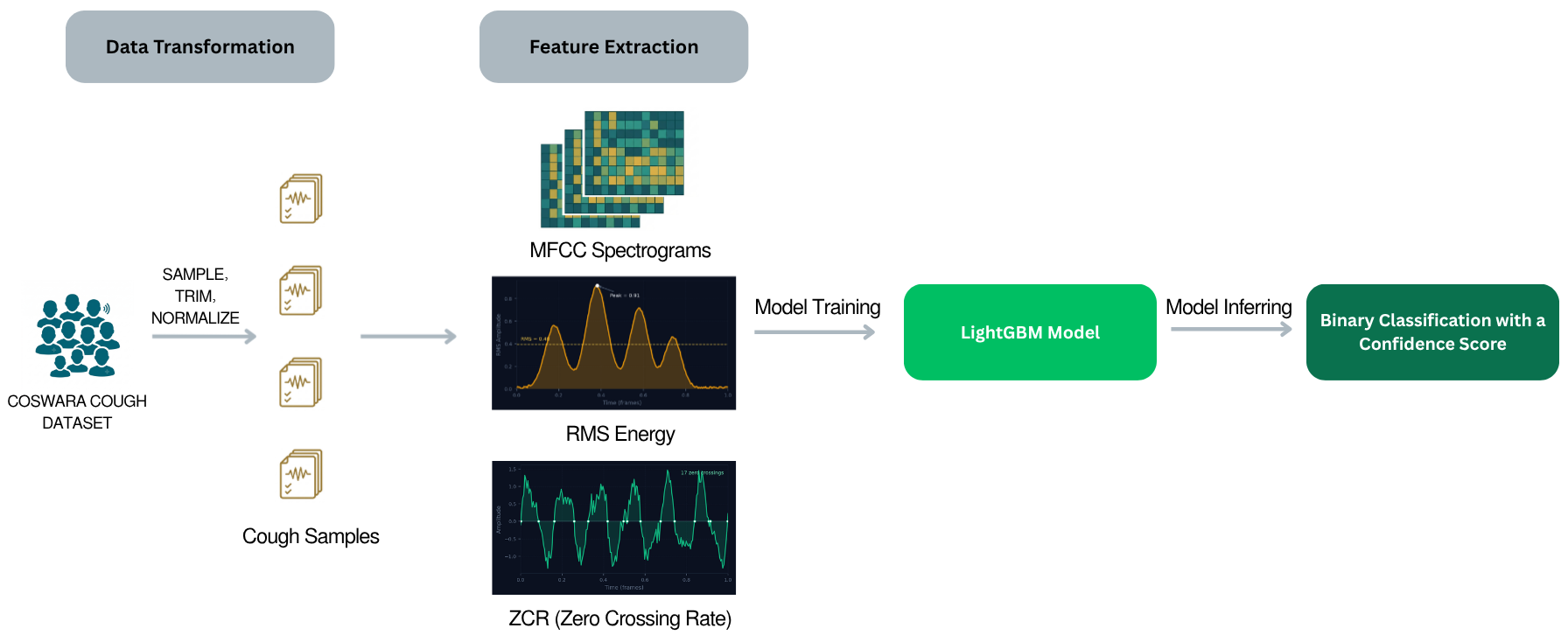}
  \caption{Cough audio processing pipeline within the MultiSense-Pneumo multimodal
framework. Pre-processed cough recordings are transformed into three acoustic
features: MFCC ($\mu_k$, $\sigma_k$), RMS energy, and ZCR, which are
concatenated into a fixed-length feature vector fed to the LightGBM classifier,
outputting a binary pneumonia prediction with a confidence score.}
  \label{fig:cough_pipeline}
\end{figure}

Figure~\ref{fig:cough_pipeline} illustrates the cough audio processing pipeline
within the broader MultiSense-Pneumo multimodal framework, from raw waveform ingestion
through feature extraction to the statistical feature vector that feeds the
LightGBM classifier. Cough recordings are drawn from the Coswara dataset
\cite{bhattacharya2023coswara}, a large-scale repository of respiratory sounds
collected for remote screening of SARS-CoV-2 infection. Specifically, only the
\texttt{cough-heavy} and \texttt{cough-shallow} audio files were used from each
participant; other recordings such as \texttt{breathing-deep} were excluded from
this analysis.

\subsubsection{Signal preprocessing}

The cough pathway treats respiratory audio as a short-duration acoustic
biomarker. Input recordings undergo three front-end operations: (i)~conversion
to monophonic waveforms, (ii)~resampling to a fixed 16\,kHz sampling rate, (iii)~segmentation into non-overlapping fixed-length windows of duration $2$ seconds, retaining only complete segments, and
(iv)~amplitude normalisation to a consistent peak level. This standardisation
is important because learned tabular and boosting-based classifiers are
sensitive to mismatches in feature extraction procedures; ensuring uniform
sampling rate and amplitude scale prevents spurious feature shifts between
training and inference.

\subsubsection{Time--frequency representation}

For a discrete cough waveform $x[n]$, the system extracts mel-frequency cepstral
coefficients (MFCCs), which provide a compact approximation of the short-term
spectral envelope of the respiratory signal. Let $M \in \mathbb{R}^{K \times T}$
denote the MFCC matrix, where $K$ is the number of cepstral coefficients and $T$
is the number of temporal frames. The feature extractor summarizes this matrix
using first- and second-order temporal statistics:
\begin{equation}
\mu_k = \frac{1}{T}\sum_{t=1}^{T} M_{k,t},
\qquad
\sigma_k = \sqrt{\frac{1}{T}\sum_{t=1}^{T}\left(M_{k,t}-\mu_k\right)^2},
\end{equation}
for $k=1,\ldots,K$. Here $\mu_k$ captures the mean spectral energy in the
$k$-th cepstral band across the recording, while $\sigma_k$ quantifies
within-signal temporal variability in that band. Together, these statistics
encode the average spectral profile of the cough event and the degree to which
energy in each band fluctuates over time---both of which may differ
systematically between pneumonia-positive and healthy respiratory patterns.

\begin{figure}[htbp]
  \centering
  \begin{subfigure}[b]{0.48\linewidth}
    \centering
    \includegraphics[width=\linewidth]{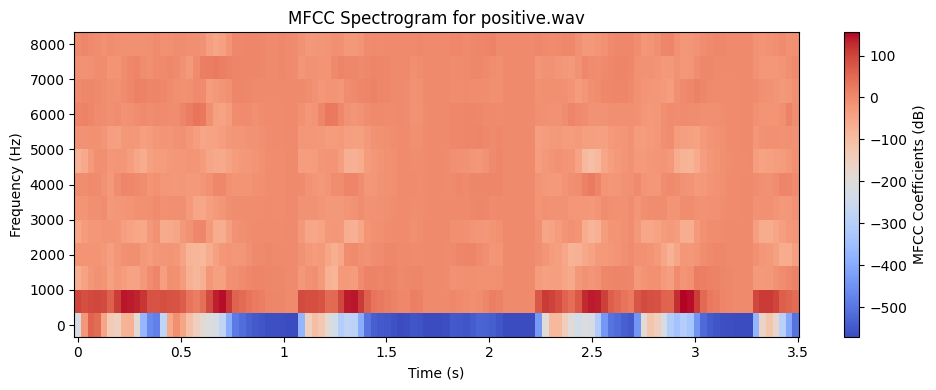}
    \caption{MFCC spectrogram of a \textbf{pneumonia-positive} cough sample.}
    \label{fig:mfcc_positive}
  \end{subfigure}
  \hfill
  \begin{subfigure}[b]{0.48\linewidth}
    \centering
    \includegraphics[width=\linewidth]{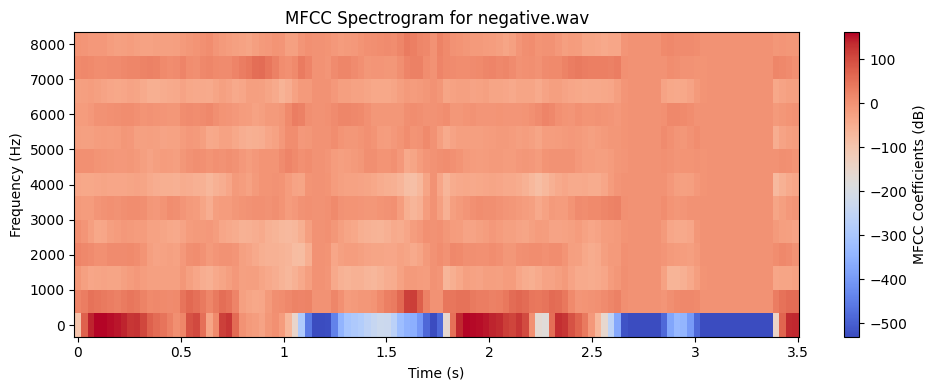}
    \caption{MFCC spectrogram of a \textbf{pneumonia-negative} cough sample.}
    \label{fig:mfcc_negative}
  \end{subfigure}
  \caption{MFCC spectrograms ($K$ coefficients $\times$ $T$ frames) for
  representative cough recordings. Warmer tones indicate higher coefficient
  magnitude. (a)~The positive sample shows elevated energy in lower cepstral
  bands and greater temporal variability; (b)~the negative sample exhibits
  a more uniform energy distribution, consistent with unobstructed airflow.}
  \label{fig:mfcc_comparison}
\end{figure}

Figure~\ref{fig:mfcc_comparison} shows representative MFCC spectrograms for
pneumonia-positive and pneumonia-negative cough recordings. Differences in
coefficient magnitude and temporal energy distribution across the two cases
motivate the use of MFCC-based features as a discriminative acoustic
representation within the cough pathway.

% \subsubsection{Auxiliary acoustic descriptors}

% In addition to MFCC summaries, the cough branch computes three low-level
% descriptors that characterise complementary acoustic properties of the
% respiratory signal:
% \begin{itemize}[leftmargin=*]
% \item \textbf{Root-mean-square (RMS) energy} --- measures the frame-level
% signal power, capturing the loudness and burst intensity of each cough event.
% Higher RMS values are associated with forceful expulsion, which may reflect
% airway inflammation or fluid accumulation.
% \item \textbf{Zero-crossing rate (ZCR)} --- counts the rate at which the
% waveform changes sign per frame, reflecting temporal irregularity and airflow
% turbulence. Wet, fluid-damped coughs tend to exhibit lower ZCR than dry or
% turbulent coughs.
% \item \textbf{Spectral centroid} --- computes the amplitude-weighted centre of
% mass of the frequency spectrum, serving as a measure of spectral brightness.
% Shifts in the centroid may indicate changes in airway patency associated with
% pneumonic consolidation.
% \end{itemize}
% The mean and standard deviation of each descriptor are computed across frames
% and concatenated with the MFCC-derived statistics, yielding a fixed-dimensional
% tabular feature vector
% \begin{equation}
% \mathbf{z} \in \mathbb{R}^{d},
% \end{equation}
% where $d$ corresponds to the total number of summary statistics produced by the
% acoustic front end.

\subsubsection{Auxiliary acoustic descriptors}

In addition to MFCC summaries and their derivatives ($\Delta, \Delta^2$), the cough branch computes six low-level descriptors (LLDs) that characterize complementary acoustic and physical properties of the respiratory signal:

\begin{itemize}[leftmargin=*]
    \item \textbf{Root-mean-square (RMS) energy} --- measures frame-level signal power, capturing the loudness and burst intensity of each cough event. Higher RMS values are associated with forceful expulsion, which may reflect airway inflammation or fluid accumulation.
    \item \textbf{Zero-crossing rate (ZCR)} --- counts the rate at which the waveform changes sign per frame, reflecting temporal irregularity and airflow turbulence. Wet, fluid-damped coughs tend to exhibit lower ZCR than dry or turbulent coughs.
    \item \textbf{Spectral Centroid} --- computes the amplitude-weighted center of mass of the frequency spectrum, serving as a measure of spectral brightness. Shifts in the centroid may indicate changes in airway patency associated with pneumonic consolidation.
    \item \textbf{Spectral Bandwidth} --- quantifies the "spread" of the spectrum around the centroid, helping to distinguish between tonal respiratory sounds and noisy, wide-band cough bursts.
    \item \textbf{Spectral Rolloff} --- identifies the frequency below which a specified percentage (e.g., 85\%) of the total spectral energy lies, providing a measure of spectral shape and high-frequency content.
    \item \textbf{Spectral Flatness} --- quantifies how much the spectrum resembles a pure tone versus white noise, providing an estimate of the "noisiness" or turbulence of the cough.
\end{itemize}

For each of these six descriptors, the system computes eight summary statistics: the mean, standard deviation, minimum, maximum, 25th and 75th percentiles, skewness, and kurtosis. These are concatenated with the MFCC-derived statistics to yield a fixed-dimensional tabular feature vector 
\begin{equation}
\mathbf{z} \in \mathbb{R}^{d},
\end{equation}
where $d$ corresponds to the total number of summary statistics produced by the acoustic front end.

% \subsubsection{LightGBM classifier}

% The acoustic classifier is a gradient-boosted decision-tree ensemble implemented in the LightGBM family. Let $y \in \{0,1\}$ denote the binary cough label, with $y=1$ corresponding to the positive or abnormal respiratory-cough class in the training corpus. Given the feature vector $\mathbf{z}$, the model estimates the posterior probability
% \begin{equation}
% \hat{\pi}_1(\mathbf{z}) = P(y=1 \mid \mathbf{z}).
% \end{equation}
% This posterior is used as the cough-related signal
% \begin{equation}
% \hat{s}_{\mathrm{cgh}} = \hat{\pi}_1(\mathbf{z}),
% \end{equation}
% or, more generally, as a monotone measure of acoustic concern aligned with the abnormal class.

% The attraction of this pathway is that it remains lightweight and interpretable at the feature level. Its main weakness, as shown in the results section, is substantial sensitivity to class imbalance and low minority-class recall.

\subsubsection{LightGBM classifier}

The acoustic classifier is a gradient-boosted decision-tree ensemble implemented using the LightGBM framework. Given the clinical nature of pneumonia screening, where healthy samples typically outnumber positive cases, the model is trained using a \textbf{stratified 5-fold cross-validation} strategy. This ensures that the proportion of pneumonia-positive samples remains consistent across training and validation splits, allowing for a reliable estimate of generalization performance.

To address class imbalance, we employ two complementary weighting strategies during the training phase. First, we set the \textit{scale\_pos\_weight} to the ratio of negative to positive samples ($N_{neg} / N_{pos}$). Second, an explicit \textit{class\_weight} dictionary is used to assign a higher importance (3:1) to the minority pneumonia class. The optimization objective is defined as binary log-loss, while the Area Under the Receiver Operating Characteristic Curve (\textbf{AUC-ROC}) is utilized as the primary evaluation metric to guide early stopping.

Let $\mathbf{z}$ denote the concatenated acoustic feature vector. The model estimates the posterior probability
\begin{equation}
\hat{\pi}_1(\mathbf{z}) = P(y=1 \mid \mathbf{z}),
\end{equation}
where $y=1$ represents the pneumonia-positive class. This probability serves as the acoustic risk signal
\begin{equation}
\hat{s}_{\mathrm{cgh}} = \hat{\pi}_1(\mathbf{z}).
\end{equation}

By utilizing $L_1$ and $L_2$ regularization ($\alpha=0.1, \lambda=0.1$) alongside early stopping, the classifier maintains robustness against overfitting on the high-dimensional statistical features while prioritizing the sensitivity required for clinical screening.

\subsection{Chest Radiograph Classification with Domain-Adversarial Learning}
\label{sec:chest_radiograph_dann}

The chest radiograph pathway serves as the primary imaging-based component of the MultiSense-Pneumo framework, providing high-resolution structural information for pneumonia assessment. This module is designed to capture pathology-relevant features while maintaining robustness to variations in image acquisition conditions. In this section, we describe the backbone architecture, the motivation for incorporating domain-adversarial learning, the construction of synthetic domain variations, and the corresponding training objective and evaluation strategy.

\subsubsection{Backbone architecture}

The imaging pathway employs a ResNet-18 convolutional neural network adapted for \emph{binary} classification of pneumonia versus normal cases. ResNet-18 provides an effective balance between representational capacity and computational efficiency, while residual skip connections facilitate stable optimization and improved gradient flow.

The final fully connected layer is modified to produce logits for two output classes. Depending on preprocessing configuration, input radiographs may be represented as either \emph{single-channel (grayscale)} or \emph{three-channel (RGB)} images. In the grayscale setting, when initializing from ImageNet-pretrained weights, the first convolutional layer can be adapted by averaging the pretrained RGB filters across channels. All images are resized to a fixed spatial resolution of $224 \times 224$, normalized, and processed through the network backbone. In this work, experiments are conducted using the \textbf{PneumoniaMNIST}~\cite{kermany2018identifying} dataset from the MedMNIST~\cite{yang2021medmnist} collection, which consists of standardized chest radiograph patches.

\subsubsection{Motivation for domain adaptation}

A major challenge in radiograph-based pneumonia classification is \emph{distribution shift}. Even when diagnostic labels remain consistent, the visual characteristics of chest X-rays can vary significantly due to factors such as blur, sensor noise, exposure differences, contrast variations, and acquisition protocols. Models trained on a narrow distribution may inadvertently learn spurious correlations tied to these domain-specific characteristics rather than pathology-relevant features.

To address this issue, the imaging pathway incorporates \emph{domain-adversarial training}, which encourages the learned feature representations to remain discriminative for pneumonia classification while being invariant to domain-specific variations.

\subsubsection{Synthetic domain construction}

In the absence of large-scale, explicitly domain-labeled multi-institution datasets, domain variation is approximated through a controlled synthetic construction. Each input image is associated with a domain label $d$ generated by applying one of several transformations:
\begin{itemize}
\item \textbf{Clean:} no transformation,
\item \textbf{Gaussian blur:} simulating loss of sharpness,
\item \textbf{Additive Gaussian noise:} simulating sensor noise,
\item \textbf{Contrast scaling:} simulating variations in exposure and intensity.
\end{itemize}

This process yields tuples $(x, y, d)$, where $x$ denotes the image, $y$ the pneumonia label, and $d$ the domain index. While synthetic domains do not fully capture real-world heterogeneity across institutions and devices, they provide a reproducible setting for evaluating robustness to controlled perturbations.

\begin{figure}[htbp]
\centering
\includegraphics[width=\textwidth]{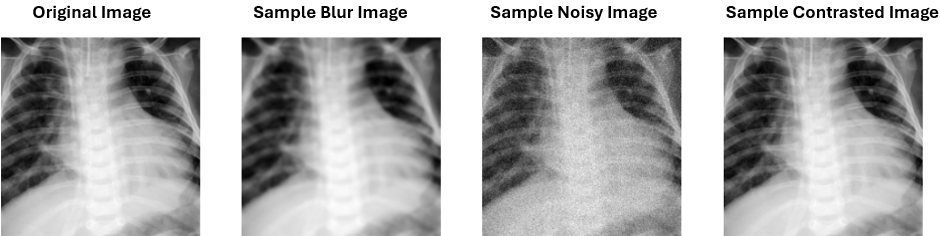}
\caption{
Examples of synthetic domain perturbations applied to chest radiographs for domain-adversarial training. From left to right: original image, Gaussian blur, additive Gaussian noise, and contrast scaling. These transformations simulate variations in imaging conditions such as blur, sensor noise, and exposure differences. Training with such perturbations encourages the model to learn pathology-relevant features while remaining robust to domain-specific artifacts.
}
\label{fig:domain_perturbations}
\end{figure}

\subsubsection{Domain-adversarial objective}

Let $G_{\theta}$ denote the feature extractor, $C_{\phi}$ the pneumonia classifier, and $D_{\psi}$ the domain discriminator. The model is trained using a domain-adversarial objective of the form:
\begin{equation}
\mathcal{L}
=
\mathcal{L}_{y}\bigl(C_{\phi}(G_{\theta}(x)),\, y\bigr)
+
\lambda \,
\mathcal{L}_{d}\bigl(D_{\psi}(\mathrm{RevGrad}(G_{\theta}(x))),\, d\bigr)
\end{equation}
where $\mathcal{L}_{y}$ is the classification loss, $\mathcal{L}_{d}$ is the domain classification loss, $\lambda$ controls the strength of domain invariance, and $\mathrm{RevGrad}(\cdot)$ denotes a gradient reversal operation.

The gradient reversal layer acts as the identity during the forward pass but reverses the sign of gradients during backpropagation. This mechanism encourages $G_{\theta}$ to learn feature representations that are predictive for pneumonia classification while being uninformative for domain discrimination.

At inference time, only the composition $C_{\phi} \circ G_{\theta}$ is used. The imaging signal is defined as the predicted probability of the pneumonia class:
\begin{equation}
\hat{s}_{\mathrm{img}} = P(\text{pneumonia} \mid x),
\end{equation}
which serves as a calibrated scalar input to the multimodal fusion stage.

\subsubsection{Evaluation metrics}

Model performance is evaluated using accuracy, macro-averaged F1 score, area under the receiver operating characteristic curve (AUROC), and expected calibration error (ECE). All metrics are computed on the held-out split after applying a softmax to the two class logits.

\textbf{Accuracy} is the fraction of examples for which the argmax-predicted label equals the ground-truth label. \textbf{Macro-F1} averages the F1 score computed independently for each class (with equal weight per class), which better reflects behavior under class imbalance than accuracy alone. \textbf{AUROC} uses the predicted probability of the positive (pneumonia) class as a continuous score; it equals the probability that a randomly chosen positive example is ranked above a randomly chosen negative, and does not require fixing a single decision threshold. \textbf{ECE} assesses calibration using \emph{binned} reliability: predictions are grouped by their top-class confidence (maximum softmax probability), and within each bin the average confidence is compared to the empirical accuracy; ECE is the sample-weighted mean absolute gap across bins (15 equal-width bins on $[0,1]$ in our implementation).

Together, accuracy and macro-F1 summarize discrete classification quality, AUROC summarizes ranking separability across thresholds, and ECE summarizes how trustworthy the predicted probabilities are as estimates of correctness.

\subsection{Speech Transcription and Speech-Derived Clinical Signal}
\label{sec:speech_transcription_signal}

The speech pathway captures verbal clinical information that may not be explicitly provided through structured questionnaires, such as free-form narration from patients, caregivers, or community health workers. In many real-world screening scenarios, especially in low-resource settings, spoken input serves as a primary mode of data collection, making it a valuable complementary modality.

Speech signals are first transcribed into text using a pretrained automatic speech recognition (ASR) model, specifically \textbf{OpenAI Whisper Small}~\cite{radford2023robust}. The transcription process converts raw audio into a textual representation that preserves the semantic content of the spoken input while enabling downstream processing using text-based methods. Importantly, the ASR system performs transcription only and does not directly infer clinical meaning.

The resulting transcript is then used to derive a secondary clinical signal. Rather than employing complex natural language understanding models, a lightweight and interpretable approach is adopted. Specifically, the transcript is analyzed using a fixed vocabulary of clinically relevant keywords associated with pneumonia, including terms related to fever, cough, breathlessness, and chest discomfort. The presence and frequency of these terms are used to compute a bounded scalar signal representing speech-derived clinical concern.

In addition to contributing to the multimodal fusion stage, the transcript is retained as a contextual input for downstream summarization. This allows the system to incorporate patient-reported information into generated reports while maintaining traceability to the original input sources.

This design improves flexibility under realistic data-entry conditions, enabling natural language input to complement structured and sensor-derived modalities. However, the pathway inherits limitations inherent to speech recognition systems. Variability in accent, background noise, recording quality, and the use of medical terminology may introduce transcription errors, including substitutions, omissions, or semantically plausible inaccuracies. These errors can propagate into downstream processing, and therefore the speech pathway should be interpreted as an assistive transcription mechanism rather than a verified clinical record.

\subsection{Multimodal Fusion}
\label{sec:multimodal_fusion}

The final screening estimate is obtained by integrating four modality-specific signals into a unified score. Each modality produces a scalar value in the range $[0,1]$, representing pneumonia-related concern.

Let $\hat{s}_{\mathrm{img}}$ denote the imaging-based signal derived from the chest radiograph classifier, expressed as the predicted probability of the pneumonia class. Let $\hat{s}_{\mathrm{sym}}$ represent the structured symptom signal obtained from questionnaire-derived clinical features. Let $\hat{s}_{\mathrm{sp}}$ denote the speech-derived signal computed from the transcribed audio using a clinically relevant keyword set, and let $\hat{s}_{\mathrm{cgh}}$ represent the cough-based signal derived from acoustic analysis.

These modality-specific signals are combined through a weighted linear fusion:
\begin{equation}
S = \sum_{m \in \{\mathrm{img}, \mathrm{sym}, \mathrm{cgh}, \mathrm{sp}\}} 
w_m \hat{s}_m,
\qquad
\sum_{m} w_m = 1.
\end{equation}

The proposed framework adopts a \emph{late} or post-hoc fusion strategy rather than end-to-end multimodal learning. This design choice is motivated by the absence of a fully paired multimodal patient dataset containing aligned symptom descriptions, cough audio, speech recordings, and chest radiographs from the same individuals. Late fusion enables modular analysis, allows each component to be trained on domain-specific data, and supports missing-modality operation. However, it also means that the fused score is a heuristic screening estimate rather than an empirically validated improvement over any single modality.

In this prototype, fixed weights are assigned as
\begin{equation}
w_{\mathrm{img}} = 0.40,\qquad
w_{\mathrm{sym}} = 0.20,\qquad
w_{\mathrm{cgh}} = 0.20,\qquad
w_{\mathrm{sp}} = 0.20.
\end{equation}

This configuration assigns the highest weight to imaging because the radiograph pathway has the strongest component-level evidence in our experiments, while symptoms, cough, and speech provide complementary contextual signals. These weights are not learned from paired multimodal outcomes and are not claimed to be statistically optimal or clinically calibrated. They should therefore be interpreted as transparent heuristic weights chosen for prototype-level aggregation and inspection.

The fused score $S$ is constrained to the interval $[0,1]$ and mapped to ordinal risk categories using fixed thresholds:
\[
\texttt{HIGH} \text{ if } S \ge 0.75, \quad
\texttt{MODERATE} \text{ if } 0.50 \le S < 0.75, \quad
\texttt{LOW} \text{ if } S < 0.50.
\]

\paragraph{Sensitivity of the heuristic fusion operator.}
Because the weights are hand-specified, we examine the sensitivity of the fusion rule at the operator level. If a weight shift $\delta$ is moved from modality $a$ to modality $b$ while preserving $\sum_m w_m=1$, the fused score changes by
\begin{equation}
\Delta S = \delta(\hat{s}_b-\hat{s}_a),
\qquad
|\Delta S| \leq \delta.
\end{equation}
Thus, for a moderate weight perturbation of $\delta=0.10$, the fused score can change by at most $0.10$ in the worst case. A risk-band assignment can therefore change only when the original score lies close to a threshold, unless the modality signals are strongly discordant. Table~\ref{tab:fusion_weight_sensitivity} summarizes representative weight configurations used to inspect this behavior. This analysis does not replace paired-patient validation; it only characterizes how the deterministic operator behaves under plausible weight choices.

\begin{table}[htbp]
\centering
\caption{Representative fusion-weight configurations for heuristic sensitivity inspection. These settings are not learned from paired multimodal data and are used only to characterize the stability of the fusion operator.}
\label{tab:fusion_weight_sensitivity}
\setlength{\tabcolsep}{6pt}
\renewcommand{\arraystretch}{1.15}
\begin{tabular}{lcccc}
\toprule
\textbf{Configuration} & $w_{\mathrm{img}}$ & $w_{\mathrm{sym}}$ & $w_{\mathrm{cgh}}$ & $w_{\mathrm{sp}}$ \\
\midrule
Base setting & 0.40 & 0.20 & 0.20 & 0.20 \\
Image-dominant & 0.55 & 0.15 & 0.15 & 0.15 \\
Cough-downweighted & 0.45 & 0.25 & 0.10 & 0.20 \\
Symptom-dominant & 0.30 & 0.35 & 0.15 & 0.20 \\
Balanced non-image & 0.25 & 0.25 & 0.25 & 0.25 \\
\bottomrule
\end{tabular}
\end{table}

The purpose of this fusion module is interpretability and modularity, not proof of clinical superiority. In particular, this study does not claim that the fused score outperforms image-only screening. Demonstrating that claim would require a paired multimodal dataset, prospective evaluation, and clinically meaningful outcome measures. Therefore, the hand-specified fusion weights should not be interpreted as clinically final. In real-world use, modality weights, risk thresholds, and triage categories should be determined through clinician-guided calibration, preferably with input from multiple medical experts, and then validated within the target deployment setting before clinical use.

\subsection{Narrative Report Generation and Multilingual Localization}
\label{sec:narrative_report_localization}

\begin{figure*}[htbp]
\centering
\includegraphics[width=0.85\linewidth]{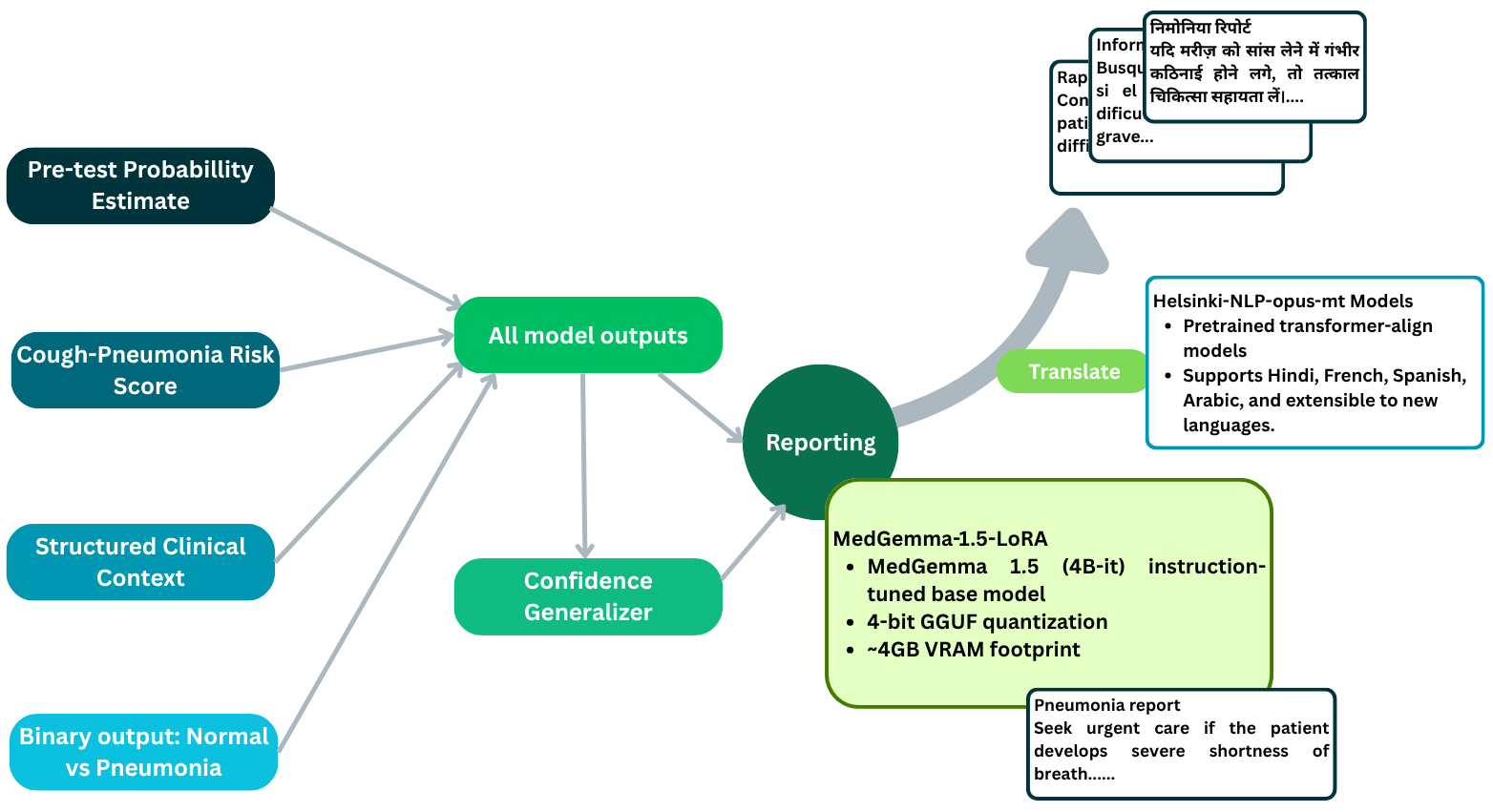}
\caption{
Overview of the MultiSense-Pneumo multimodal pipeline. Modality-specific inputs, including clinical, acoustic, and imaging signals, are aggregated and processed to produce a unified screening score. The fused output can be used to generate a structured research summary and optional translations, but these language outputs are not clinically validated reports.
}
\label{fig:MultiSense-Pneumo_pipeline}
\end{figure*}

Beyond scalar prediction, the framework includes optional language-based components for generating human-readable summaries and enabling multilingual accessibility. These components are designed to make the system outputs more transparent, interpretable, and easier to communicate in research demonstrations and potential decision-support workflows (Fig.~\ref{fig:MultiSense-Pneumo_pipeline}).

\paragraph{Report synthesis.}
A narrative clinical-style summary is generated using a pretrained medical language model, specifically \textbf{MedGemma 4B Instruction-Tuned} (quantized GGUF format)~\cite{sellergren2025medgemma}. The model is deployed locally through an OpenAI-compatible inference interface and operates entirely offline, preserving the framework's low-resource and privacy-aware deployment objective.

The report-generation process takes as input structured evidence aggregated from the upstream modules, including symptom descriptions, cough-related context, speech-derived transcripts, textual summaries of imaging predictions, and the integrated screening score. The model is prompted to summarize only the provided evidence and to avoid introducing unsupported clinical findings, particularly for radiographic observations. The generated report is therefore intended as a structured explanation of the prototype outputs rather than as an independent diagnostic interpretation.

\paragraph{Quality and safety considerations.}
The narrative report is intended to support communication of model outputs, not to replace clinical judgment. In particular, the report should preserve uncertainty by explicitly indicating when modalities are missing, weak, or discordant. This is important because the generated text depends on the quality of the upstream modality-specific predictions and the evidence provided to the language model. For real-world deployment, the wording, modality emphasis, risk thresholds, and reporting format should be reviewed and calibrated with qualified clinical personnel, preferably involving multiple medical experts and the intended clinical workflow.

\paragraph{Multilingual localization.}
To improve accessibility across diverse populations, the generated summary can be translated into multiple target languages using pretrained neural machine translation models. Specifically, this work employs the \textbf{Helsinki-NLP OPUS-MT}~\cite{tiedemann2020opus} models for English-to-target translation. These models are applied to the generated narrative while preserving structured risk labels and modality-specific evidence where possible. This design supports multilingual research demonstrations and can help communicate screening outputs to users in different language settings.

\paragraph{Limitations.}
The language-based components are assistive modules within the broader prototype. Generated reports may reflect errors propagated from upstream modalities, and language models may occasionally introduce omissions, unsupported details, or overly certain phrasing. Similarly, machine translation may reduce fidelity in complex medical descriptions or alter clinical terminology. Therefore, these outputs should be interpreted as explanatory summaries of the system evidence and should be reviewed by qualified personnel before any real-world clinical use.

\section{Quantitative Component Results}
\label{sec:quantitative_results}

This section presents quantitative evaluation results for the two primary classification components of MultiSense-Pneumo: the domain-adversarial radiograph classifier and the cough acoustic classifier. The current evaluation is component-level. We do not claim end-to-end clinical validation of the complete multimodal pipeline because the experiments do not use a paired patient dataset containing symptoms, cough audio, speech, and chest radiographs from the same individuals.

\subsection{Radiograph classifier under synthetic domain shifts}
\label{sec:results_radiograph}

Table~\ref{tab:radiograph_aggregate_results} presents the aggregate validation and test performance of the domain-adversarial ResNet-18 radiograph classifier. The results demonstrate strong discriminative performance and good probabilistic calibration, particularly on the validation split. While some degradation is observed on the test set, the decline remains moderate, indicating that the model generalizes effectively under the defined perturbation regime.

\begin{table}[htbp]
\centering
\caption{Aggregate validation and test performance of the domain-adversarial radiograph classifier.}
\label{tab:radiograph_aggregate_results}
\setlength{\tabcolsep}{6pt}
\renewcommand{\arraystretch}{1.2}
\begin{tabular}{lccccc}
\toprule
\textbf{Split} & \textbf{Loss} & \textbf{Accuracy} & \textbf{F1 Score} & \textbf{AUROC} & \textbf{ECE} \\
\midrule
Validation & 0.1040 & 0.9733 & 0.9654 & 0.9919 & 0.0138 \\
Test       & 0.3484 & 0.9255 & 0.9171 & 0.9752 & 0.0511 \\
\bottomrule
\end{tabular}
\end{table}

To further assess robustness, Tables~\ref{tab:radiograph_val_domain_results} and \ref{tab:radiograph_test_domain_results} report performance across synthetic domain conditions, including clean, blur, noise, and contrast perturbations.

\begin{table}[htbp]
\centering
\caption{Per-domain validation performance of the radiograph classifier.}
\label{tab:radiograph_val_domain_results}
\setlength{\tabcolsep}{6pt}
\renewcommand{\arraystretch}{1.2}
\begin{tabular}{clcccc}
\toprule
\textbf{$d$} & \textbf{Domain} & \textbf{Accuracy} & \textbf{Macro-F1} & \textbf{AUROC} & \textbf{ECE} \\
\midrule
0 & Clean     & 0.9733 & 0.9656 & 0.9923 & 0.0168 \\
1 & Blur      & 0.9733 & 0.9651 & 0.9922 & 0.0159 \\
2 & Noise     & 0.9790 & 0.9728 & 0.9920 & 0.0164 \\
3 & Contrast  & 0.9676 & 0.9583 & 0.9914 & 0.0196 \\
\bottomrule
\end{tabular}
\end{table}

\begin{table}[htbp]
\centering
\caption{Per-domain test performance of the radiograph classifier.}
\label{tab:radiograph_test_domain_results}
\setlength{\tabcolsep}{6pt}
\renewcommand{\arraystretch}{1.2}
\begin{tabular}{clcccc}
\toprule
\textbf{$d$} & \textbf{Domain} & \textbf{Accuracy} & \textbf{Macro-F1} & \textbf{AUROC} & \textbf{ECE} \\
\midrule
0 & Clean     & 0.9343 & 0.9275 & 0.9781 & 0.0484 \\
1 & Blur      & 0.9167 & 0.9067 & 0.9727 & 0.0572 \\
2 & Noise     & 0.9199 & 0.9103 & 0.9738 & 0.0593 \\
3 & Contrast  & 0.9311 & 0.9238 & 0.9765 & 0.0537 \\
\bottomrule
\end{tabular}
\end{table}

The relatively small variation in AUROC across domain conditions indicates that the domain-adversarial objective effectively reduces sensitivity to synthetic perturbations. Although minor degradation is observed under blur and noise, overall performance remains consistently high. This suggests that the model learns representations that are largely invariant to low-level acquisition artifacts while preserving pathology-relevant information.

Among the evaluated modality-specific components, the radiograph pathway demonstrates the strongest component-level performance. In the prototype fusion rule, it is therefore assigned the largest weight, although this does not establish clinical superiority of the fused system.

\subsection{Cough acoustic classifier}
\label{sec:results_cough}

Table~\ref{tab:cough_classifier_results} reports evaluation results for the LightGBM-based cough classifier. The results exhibit a clear class-dependent performance imbalance. The model achieves stronger performance for the normal class but substantially weaker recall for the abnormal class. In a screening context, this limitation is important because missed abnormal cases are more safety-critical than false alarms.

\begin{table}[htbp]
\centering
\caption{Evaluation of the cough acoustic classifier.}
\label{tab:cough_classifier_results}
\begin{tabular}{@{}lrrrr@{}}
\toprule
 & \textbf{Precision} & \textbf{Recall} & \textbf{F1-score} & \textbf{Support} \\ \midrule
normal (0)   & 0.86 & 0.88 & 0.87 & 452 \\
abnormal (1) & 0.44 & 0.39 & 0.41 & 109 \\ \midrule
\textbf{Accuracy}     &      &      & 0.78 & 561 \\
\textbf{Macro avg}    & 0.65 & 0.63 & 0.64 & 561 \\
\textbf{Weighted avg} & 0.77 & 0.78 & 0.78 & 561 \\ \bottomrule
\end{tabular}
\end{table}

The reduced abnormal-class recall is likely driven by a combination of limited training data, class imbalance, recording variability, and the indirect relationship between cough acoustics and radiographic pneumonia status. In this work, cough classification relies on publicly available datasets such as Coswara~\cite{bhattacharya2023coswara}, which were not collected as paired pneumonia-screening cohorts and may not fully represent the acoustic variability encountered in clinical triage.

More fundamentally, cough acoustics are a weaker and less specific signal than imaging. Microphone quality, distance from the device, background noise, cough effort, co-existing respiratory conditions, and inter-individual variability can all alter the acoustic representation. As a result, cough-based classification alone is insufficient for reliable pneumonia screening in the current prototype.

\paragraph{Safety implications for multimodal use.}
The cough pathway should be treated as an auxiliary source of evidence and should not be used to rule out pneumonia when other indicators suggest risk. In particular, a low cough score should not override severe symptoms, concerning speech-derived descriptions, or positive radiographic evidence. In the current fusion design, the cough branch contributes only a limited fraction of the final score, and its limitations motivate future work on larger, clinically paired, and more balanced respiratory audio datasets. Until such validation is available, this component should be interpreted as exploratory rather than clinically reliable.

\section{Limitations and Future Work}
\label{sec:limitations_future_work}

This study presents a modular offline framework for multimodal pneumonia screening, with emphasis on feasibility, interpretability, and low-resource deployment. A key scope boundary is that the present experiments do not use a fully paired multimodal patient-level dataset containing symptoms, cough audio, speech recordings, and chest radiographs from the same individuals. As a result, the current evaluation supports component-level performance and overall framework feasibility, while future paired-cohort studies are needed to quantify whether the fused score improves screening outcomes relative to image-only models or standard clinical workflows.

The fusion weights are hand-specified to preserve transparency and make the contribution of each modality easy to inspect. This design is suitable for an interpretable prototype, especially when paired outcome data are unavailable. The sensitivity analysis in Section~\ref{sec:multimodal_fusion} characterizes how the deterministic fusion operator behaves under alternative weights. Future work can extend this module using learned fusion, uncertainty-aware fusion, and outcome-calibrated weighting once paired multimodal cohorts become available. In practical deployment, modality weights and decision thresholds should also be calibrated with input from qualified clinical personnel and adapted to the target clinical workflow.

The cough pathway provides an additional non-invasive signal, but its abnormal-class recall remains limited in the current prototype. Therefore, cough acoustics are treated as an auxiliary modality rather than a standalone screening decision. Future improvements should include larger respiratory audio cohorts, pneumonia-specific labels, improved class-balancing strategies, sensitivity-oriented threshold tuning, and external validation across recording devices and environments.

The narrative report generator and multilingual translation modules are included to improve interpretability, communication, and accessibility of the system outputs. These components summarize structured evidence from the upstream modules rather than generating independent clinical diagnoses. Future work should include structured factuality checks, clinician review, calibration of generated statements against source evidence, and safeguards that preserve uncertainty and prevent diagnostic overstatement, especially in multilingual settings.

Finally, the offline implementation demonstrates that the proposed framework can operate under modest computational constraints, supporting its relevance for low-resource and field-oriented scenarios. Before clinical deployment, the system would require prospective validation, institutional review, privacy and security assessment, usability testing with frontline workers, and evaluation of patient-safety outcomes. These steps represent the natural progression from a research prototype toward a clinically responsible decision-support system.

\section{Conclusion}
\label{sec:conclusion}

We presented MultiSense-Pneumo, a modular multimodal framework for pneumonia-oriented screening and triage research in resource-constrained settings. The system integrates structured symptoms, cough acoustics, speech-derived cues, and chest radiographs through interpretable late fusion, with optional narrative summarization and multilingual localization. The radiograph pathway achieved strong component-level performance under controlled synthetic domain shifts, while the cough pathway revealed important limitations in abnormal-class recall. The revised framing emphasizes that MultiSense-Pneumo is a framework and component-level prototype rather than a clinically validated diagnostic system. Future work should focus on paired multimodal data collection, empirical validation of fusion strategies, safety-aware report generation, and prospective evaluation in real frontline screening environments.

\bibliographystyle{splncs04}
\bibliography{references}

\end{document}